\documentclass[pre,twocolumn,twoside,byrevtex,superscriptaddress,floatfix,nofootinbib]{revtex4-1}

\usepackage{currfile}
\usepackage{color}

\usepackage{graphicx,epsfig,verbatim,enumerate}
\usepackage{amssymb,amsmath}
\usepackage{ifthen}

\usepackage{longtable}
\usepackage[table]{xcolor}
\definecolor{lightgray}{gray}{0.7}

\usepackage{mathtools}

\newboolean{twocolswitch}

\newcommand{\sindex}[1]{}
\newcommand{\nindex}[1]{}

\newcommand{\www}[1]{\url{#1}}

\usepackage{lettrine}

\usepackage[caption=false]{subfig}

\usepackage{siunitx}

\usepackage{array}
\newcolumntype{L}[1]{>{\raggedright\let\newline\\\arraybackslash\hspace{0pt}}p{#1}}
\newcolumntype{C}[1]{>{\centering\let\newline\\\arraybackslash\hspace{0pt}}p{#1}}
\newcolumntype{R}[1]{>{\raggedleft\let\newline\\\arraybackslash\hspace{0pt}}p{#1}}

\usepackage{hyperref}
\urldef\mapsurl
\url{http://www.uvm.edu/~tgray3/Research/Verb_Regularization/Maps/Maps.html}
\urldef\demvarsurl
\url{http://www.uvm.edu/~tgray3/Research/Verb_Regularization/VerbRegularization.html}
\urldef\onlineappendixurl
\url{https://www.uvm.edu/storylab/share/papers/gray2018a/}
\urldef\ngramsfigurl\url{https://books.google.com/ngrams/graph?content=burned%2Cburnt&year_start=1800&year_end=2000&corpus=16&smoothing=3} 

\setboolean{twocolswitch}{true}
\newcommand{\revtexlatexswitch}[2]{#1}

\begin{document}

\title{\protect
English verb regularization in books and tweets
}

\author{
\firstname{Tyler J.}
\surname{Gray}
}
\email{tyler.gray@uvm.edu}

\affiliation{
  Vermont Complex Systems Center,
  Computational Story Lab,
  Department of Mathematics \& Statistics,
  The University of Vermont,
  Burlington, VT 05401.
  }

\author{
\firstname{Andrew J.}
\surname{Reagan}
}
\email{areagan@massmutual.com}

\affiliation{MassMutual Data Science, 
Amherst, MA 01002.}

\author{
\firstname{Peter Sheridan}
\surname{Dodds}
}
\email{peter.dodds@uvm.edu}

\affiliation{
  Vermont Complex Systems Center,
  Computational Story Lab,
  Department of Mathematics \& Statistics,
  The University of Vermont,
  Burlington, VT 05401.
  }

\author{
\firstname{Christopher M.}
\surname{Danforth}
}
\email{chris.danforth@uvm.edu}

\affiliation{
  Vermont Complex Systems Center,
  Computational Story Lab,
  Department of Mathematics \& Statistics,
  The University of Vermont,
  Burlington, VT 05401.
  }

\date{\today}

\begin{abstract}
  \protect
  The English language has evolved dramatically throughout its lifespan, to the extent that a modern speaker of Old English would be incomprehensible without translation. One concrete indicator of this process is the movement from irregular to regular (-ed) forms for the past tense of verbs. In this study we quantify the extent of verb regularization using two vastly disparate datasets: (1) Six years of published books scanned by Google (2003--2008), and (2) A decade of social media messages posted to Twitter (2008--2017). We find that the extent of verb regularization is greater on Twitter, taken as a whole, than in English Fiction books. Regularization is also greater for tweets geotagged in the United States relative to American English books, but the opposite is true for tweets geotagged in the United Kingdom relative to British English books. We also find interesting regional variations in regularization across counties in the United States. However, once differences in population are accounted for, we do not identify strong correlations with socio-demographic variables such as education or income.

\end{abstract}


\maketitle

\section{Introduction}
\label{sec:introduction}

Human language reflects cultural, political, and social evolution. Words are the atoms of language.  Their meanings and usage patterns reveal insight into the dynamical process by which society changes. Indeed, the increasing frequency with which electronic text is used as a means of communicating, e.g., through email, text messaging, and social media, offers us the opportunity to quantify previously unobserved mechanisms of linguistic development.

While there are many aspects of language being investigated towards an increased understanding of social and linguistic evolution \cite{celex,ngrams,nature,GreenhillE8822,Reali20172586,Ramiro201714730}, one particular area of focus has been on changes in past tense forms for English verbs \cite{celex,ngrams,nature}. These investigations have collectively demonstrated that English verbs are going through a process of regularization, where the original irregular past tense of a verb is replaced with the regular past tense, formed using the suffix -ed.

\begin{figure*}[tp!]
  \centering	
\includegraphics[width=\textwidth]{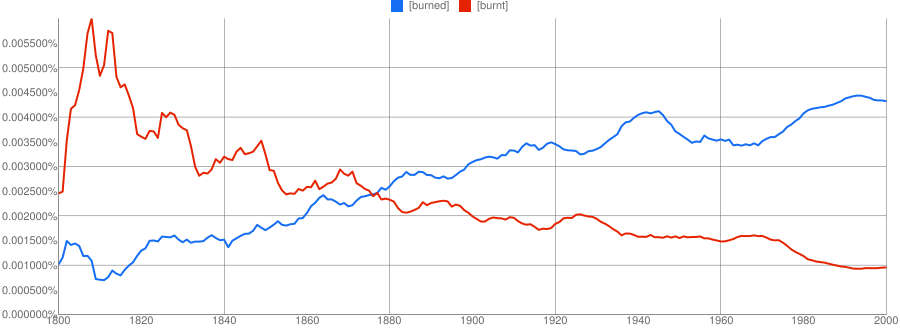}
\caption{\protect Relative word frequencies for the irregular and regular past verb forms for `burn' during the 19th and 20th centuries, using the Google Ngram Online Viewer with the English Fiction 2012 corpus. Google Ngram trends can be misleading but capture basic shifts in a language's lexicon \cite{dodds,eitan2}.  The irregular form `burnt' was once more popular, but the regular form `burned' overtook it in the late 19th century and its popularity has steadily increased ever since while that of `burnt' has decreased.  The dynamics of verb tense changes are rich, reflecting many processes at play in the Google Books Ngram data. An interactive version of this graphic can be found at
\ngramsfigurl.
}\label{burnNgrams}
\end{figure*}

For example, the irregular past tense of the verb `burn' is `burnt' and the regular past tense is `burned'. Over time, the regular past tense has become more popular in general, and for some verbs has overtaken the irregular form.  For example, in Fig.~\ref{burnNgrams}, we use the Google Ngram Online Viewer to compare the relative frequency of `burnt' with that of `burned' over the past 200 years. (As shown in an earlier paper involving two of the present authors \cite{dodds}, and expanded on below, the Google Ngram dataset is highly problematic but can serve as a useful barometer of lexical change.)   In the first half of the 19th century, the irregular past tense `burnt' was more popular.  However, the regular past tense `burned' gained in popularity and in the late 1800s became the more popular form, which has persisted through to today.

Looking at several examples like this, in a 2011 paper Michel et al.\ studied the regularization of verbs, along with other cultural and language trends, as an accompaniment to their introduction of the Google Books Ngram corpus (hereafter Ngrams) and the proto-field `Culturomics' \cite{ngrams}.  They found that most of the verb regularization over the last two centuries came from verbs using the suffix -t for the irregular form, and that British English texts were less likely than American English ones to move away from this irregular form.   

In a 2007 study, Lieberman et al.\ explored the regularization of English verbs using the CELEX corpus, which gives word frequencies from several textual sources \cite{celex}. Focusing on a set of 177 verbs that were all irregular in Old English, they examined how the rate of verb regularization relates to frequency of usage, finding that more common verbs regularized at a slower rate.  They calculated half-lives for irregular verbs binned by frequency, finding that irregular verbs regularize with a half-life proportional to the square root of frequency of usage.  

In a more recent study, Newberry et al.\ proposed a method for determining the underlying mechanisms driving language change, including the regularization of verbs \cite{nature}. Using the Corpus of Historical American English and inspired by ideas from evolution, the authors described a method to determine if language change is due to selection or drift, and applied this method to three areas of language change. They used a null hypothesis of stochastic drift and checked if selection would be strong enough to reject this null hypothesis. Of the 36 verbs Newberry et al.\ studied, only six demonstrated statistical support for selection. They also claimed that rhyming patterns might be a driver of selection.

Unfortunately, the corpora used in these studies have considerable limitations and corruptions.  
For example, early versions of the Ngrams data includes scientific literature, whose explosive growth through the 20th century is responsible for the decreasing trend in relative word usage frequency observed in many common search terms \cite{dodds}. Moreover, the library-like nature of the corpus admits no accounting for popularity: Lord of the Rings and an unknown work contribute with equal weight to token counts.

Another general concern with large corpora of a global language like English is that language use varies tremendously with culture and geography.  Ngrams allows only for the regional exploration of the English language with the British English corpus and the American English corpus.  Twitter data enables us to focus on much smaller spatial regions (e.g., county or state).

Prior studies of verb regularization have also focused on data reflecting a formal editorial process, such as the one undergone by any published book. This editorial process will tend to normalize the language, reflecting the linguistic opinions of a small minority of canon gatekeepers, rather than 
portray the language used by everyday people.  For example,  maybe the irregular from of a particular verb is considered proper by scholars, but a vast majority of the English speaking population uses the regular form. While it is not a verb form, one illustrative example is `whom'. Although `whom' is the correct word to use in the objective case, it is common for everyday speakers to use `who'. 

In the present study we take tweets to be a closer representation of everyday language.  For the vast majority of accounts, tweets are authored by individuals without undergoing a formal editing process. As such, the language therein should more accurately represent average speakers than what is found in books. 

The demographic groups contributing to Twitter are by no means a carefully selected cross-section of society, but do offer natural language use by the roughly 20\% of adult English speakers who use Twitter \cite{pew}. When exploring temporal changes in language use, the Ngrams and CELEX datasets evidently cover a much longer period than the decade for which social media is available. As a result, we are unable to infer anything about the temporal dimension of regularization looking at Twitter.

In this paper we use the Ngrams and Twitter datasets to establish estimates of the current state of English verb regularization.  We structure our paper as follows:  In Sec.~\ref{sec:data}, we describe the datasets we use.  In Sec.~\ref{sec:results}, we present our results. We study verb regularization in English in general in Sec.~\ref{sec:general_english}. We compare verb regularization in American English (AE) and British English (BE) using both Ngrams and geotagged Twitter data in Sec.~\ref{sec-BEvAE}.  In Sec.~\ref{sec:county}, we employ methods to study regional variation in verb usage, leveraging county level user location data in the United States.  We also explore correlations between verb regularization and a number of socio-demographic and economic variables.  Finally, in Sec.~\ref{sec:ConcludingRemarks}, we provide concluding remarks.  

\section{Description of data sets}
\label{sec:data}

To be consistent with prior work, we chose the verb list for our project to match that of Michel et al.\ \cite{ngrams}.  When comparing BE with AE, we use the subset of verbs that form the irregular past tense with the suffix -t.  When calculating frequencies or token counts for the `past tense' we use both the preterite and past participle of the verb. See 
\revtexlatexswitch{Table~\ref{table:verb_table} in Appendix \ref{appendix:verb_table}}{S1 Table}
 for a complete tabulation of all verb forms.

The Ngrams data reflects relative frequency, providing, for a verb and a given year, the percentage of corpus tokens that are the given verb, where a token is an individual occurrence of a word. The Google Ngram Online Viewer also has a smoothing parameter, $s$, which averages the relative frequency for the given year with that of each of the $s$ years before and after the given year, if they exist.  For example, Fig.~\ref{burnNgrams} uses a smoothing of 3 years and shows that, averaged across the years 1997--2000 (the value displayed for the year 2000), the word `burned' appeared with relative frequency 0.004321\% (roughly once every 23,000 tokens), while `burnt' appeared with relative frequency 0.000954\% (roughly once every 105,000 tokens).

We downloaded the Ngrams verb data for the most recent 6-year period available (2003--2008) \cite{ngramsViewer}. Specifically, we chose the 2008 values of relative frequency with a smoothing of 5 years, resulting in an average case insensitive\revtexlatexswitch{\footnote{
When Ngrams computes a case insensitive word frequency it uses ``the yearwise sum of the most common case-insensitive variants of the input query'' \cite{ngramsInfo}.}
word frequency for the years 2003--2008.}{ word frequency for the years 2003--2008.  (When Ngrams computes a case insensitive word frequency it uses ``the yearwise sum of the most common case-insensitive variants of the input query'' \cite{ngramsInfo}.)}
  For general English, as suggested by \cite{dodds}, we queried the English Fiction 2012 corpus, which uses ``books predominantly in the English language that a library or publisher identified as fiction.''  For AE we used the American English 2012 corpus, which uses ``books predominantly in the English language that were published in the United States.''  For BE we used the British English 2012 corpus, which uses ``books predominantly in the English language that were published in Great Britain''  \cite{ngramsInfo}.

The Twitter messages for our project consist of a random sample of roughly 10\% of all tweets posted between 9 September 2008 and 22 October 2017. This `decahose' dataset comprises a total of more than 106 billion messages, sent by about 750 million unique accounts. From this larger set, we performed a case-insensitive search for verb forms of interest, also extracting geographic location when available in the meta-data associated with each tweet. 
Tweets geotagged by mobile phone GPS with a U.S. location comprise about a 0.27\% subset of the decahose dataset; United Kingdom locations comprise about a 0.05\% subset. Many individuals provide location information, entered as free text, along with their biographical profile. We matched user specified locations of the form `city, state' to a U.S. county when possible, comprising a 2.26\% subset of the decahose dataset. Details on this matching process can be found in 
\revtexlatexswitch{Appendix \ref{appendix:location_match_info}}{S1 Appendix}.

For general English, we counted the number of tokens in the decahose dataset for each verb.  For AE, we used the tweets whose geotagged coordinates are located in the United States, and for BE we used the tweets whose geotagged coordinates are located in the United Kingdom. For the analysis of verbs by county, we used the tweets with the user entered location information.  Table~\ref{table:datasets} summarizes the datasets used for both Ngrams and Twitter.

\begin{table}[htbp!]
\begin{center}
\begin{tabular}{c|p{3.5cm}|p{3.5cm}}
 & Ngrams & Twitter \\
\hline
(I) & English Fiction 2012 corpus & All tweets \\
\hline
(II) & American English 2012 corpus & All tweets geolocated in the US \\
\hline
(III) & British English 2012 corpus & All tweets geolocated in the UK \\
\hline
(IV) & N/A & All tweets with user entered location matching `city, state' \\

\end{tabular}
\end{center}
\caption{Summary of verb datasets.}\label{table:datasets}
\end{table}

The demographic data for U.S. counties comes from the 2015 American Community Survey 5-year estimates, tables DP02--Selected Social Characteristics, DP03--Selected Economic Characteristics, DP04--Selected Housing Characteristics, and DP05--Demographic and Housing Estimates, which can be found by searching online at \url{https://factfinder.census.gov/}.  These tables comprise a total of 513 usable socio-demographic and economic variables.

We compute the \textit{regularization fraction} for a verb as the proportion of instances in which the regular form was used for the past tense of the verb.  More specifically, for Ngrams we divide the relative frequency for the regular past tense by the sum of the relative frequencies for the regular and irregular past tenses.  Similarly, for Twitter we divide the token count for the regular past tense by the sum of the token counts for both the regular and irregular past tenses. If the resulting regularization fraction is greater than $0.5$, the regular past tense is more popular and we call the verb regular. Otherwise we call the verb irregular.  

When calculating an average regularization across all verbs, we first compute the regularization fraction for each verb individually. Then we compute the average of the regularization fractions, with each verb contributing the same weight in the average, irrespective of frequency. We perform this `average of averages' to avoid swamping the contribution of less frequent verbs.

\begin{figure*}[h!bt]
\begin{center}
\includegraphics[width=\textwidth]{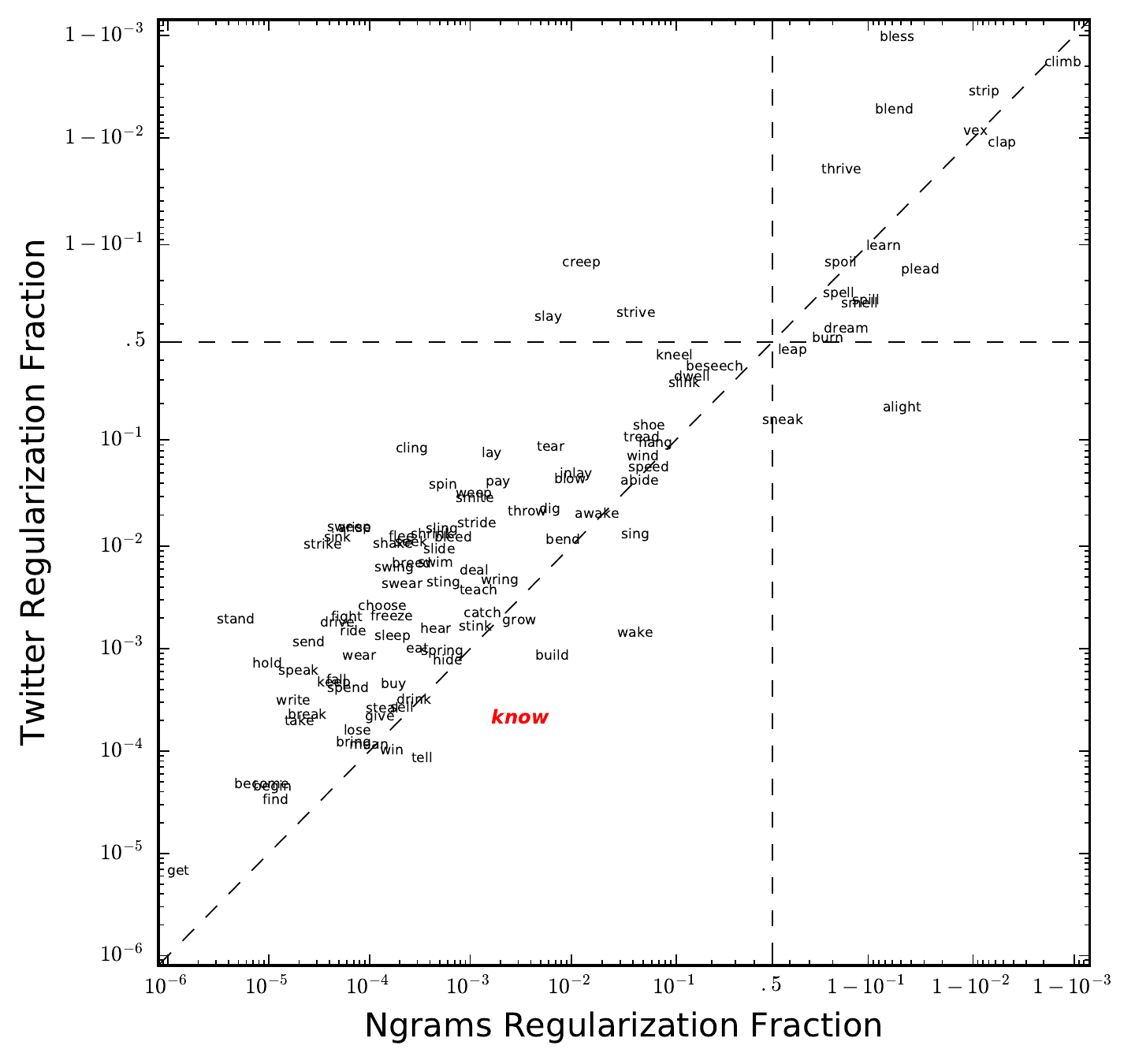}
\end{center}
\caption{Comparison of verb regularization for Ngrams and Twitter.  We calculate verb regularization fractions using the datasets in row (I) of Table~\ref{table:datasets}.  Verbs are centered at their regularization fraction in Ngrams (horizontal) and Twitter (vertical). Both axes are on a logit scale, which spreads out both extremes of the interval $(0,1)$.  Verbs to the right of the vertical dashed line are regular in Ngrams; verbs above the horizontal dashed line are regular on Twitter.  
The diagonal dashed line separates verbs that are more regular on Twitter (those above and to the left of the line) from those that are more regular in Ngrams (those below and to the right of the line).  For example, compared with `knew', the word `knowed' appears roughly 3 times in 1000 in Ngrams, and 2 times in 10,000 on Twitter, making `know' irregular in both cases, but more than an order of magnitude more regular in Ngrams than on Twitter.     }\label{ngramsVtwitter}
\end{figure*}

\begin{figure*}[h!bt]
\begin{center}
\includegraphics[width=\textwidth]{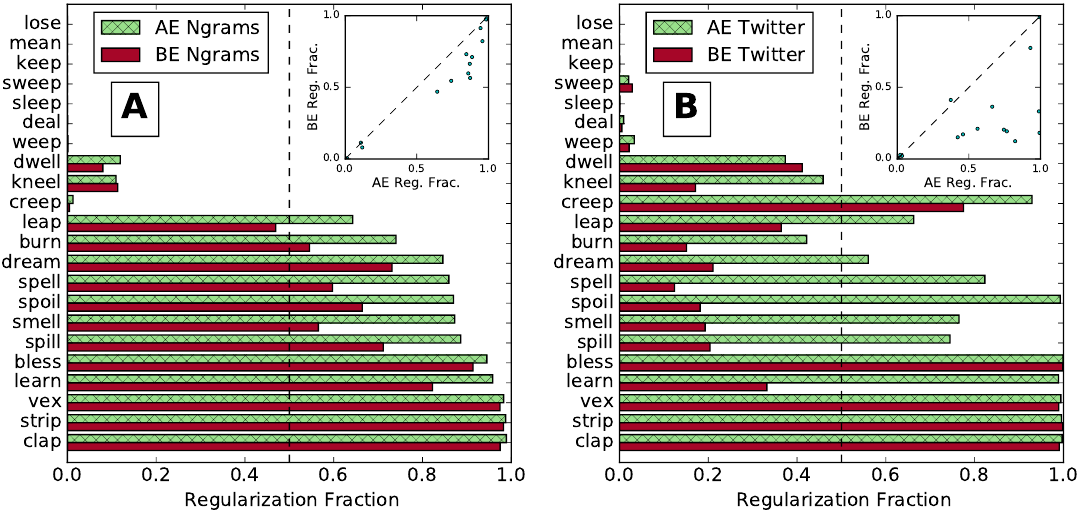}
\end{center}
\caption{American and British English verb regularization fractions for (A) Ngrams and (B) Twitter. We use the subset of verbs that form the irregular past tense with the suffix -t and the datasets in rows (II) and (III) of Table~\ref{table:datasets}. The inset scatter plot has a point for each verb. The dashed diagonal line separates verbs that are more regular in AE (below the line) from those that are more regular in BE (above the line).}\label{fig:bar_AE_BE}
\end{figure*}

\section{Methods and results}
\label{sec:results}

\subsection{Verb regularization using Ngrams and Twitter}\label{sec:general_english} 

Using the datasets in row (I) of Table~\ref{table:datasets}, we begin by comparing Ngrams and Twitter with respect to regularization of English verbs in Fig.~\ref{ngramsVtwitter}, where we find that $21$ verbs are more regular in Ngrams, and $85$ are more regular on Twitter. A Wilcoxon signed rank test of the data has a $p$-value of $7.9\times 10^{-6}$, demonstrating strong evidence that verbs on Twitter are more regular than verbs in Ngrams.  

What mechanisms could be responsible for the observed increase in
regularity on Twitter? One possibility is that authors of fiction
published in the 2000s, along with their editors, being professional
users of English, have a larger vocabulary than the typical user of
Twitter.  If so, their commitment to proper English would contribute
to the appearance of relatively more irregular verbs in books.  The
average Twitter user may not know, or choose to use, the `correct'
past tense form of particular verbs, and thus use the default regular
past tense.

Another driver may be that non-native English speakers writing English tweets may be more likely to use the default regular form. We will find quantitative support for this mechanism below.  As a preview, we note that Fig.~\ref{ngramsVtwitter} shows that `burn' is predominantly regular on Twitter globally, but we see later (Fig.~\ref{fig:bar_AE_BE}B) that `burn' is irregular on Twitter for both American English and British English. Thus, it is likely that non-native speakers are contributing to this difference.

\begin{figure*}[h!bt]
\begin{center}
\includegraphics[width=\textwidth]{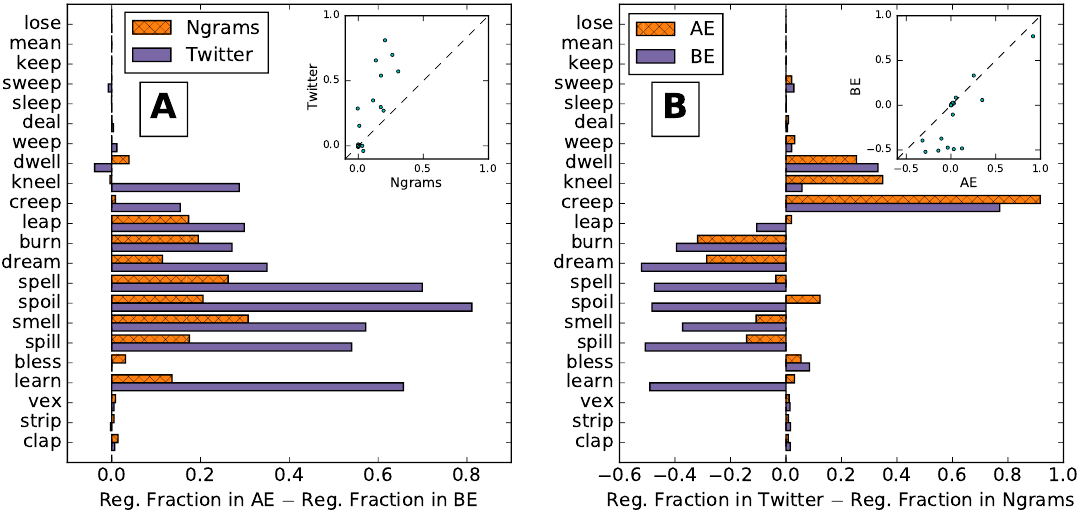}
\end{center}
\caption{Differences in verb regularization fractions.  The bar chart gives the difference for each verb in each corpus.  The inset scatter plot has a point for each verb.  (A) The difference between verb regularization fractions for AE and BE in Twitter and Ngrams.  The dashed diagonal line of the inset scatter plot separates verbs for which this difference is greater in Ngrams (below the line) from those for which it is greater in Twitter (above the line).  (B) The difference between verb regularization fraction for Twitter and Ngrams in AE and BE.  The dashed diagonal line of the inset scatter plot separates verbs for which this difference is greater in AE (below the line) from those for which it is greater in BE (above the line). }\label{fig:bar_diff}
\end{figure*}

\subsection{American and British English}\label{sec-BEvAE}

We next study how verb regularization varies with geographic region. 
In this subsection we use the datasets in row (II) of Table~\ref{table:datasets} for AE and row (III) for BE and the subset of verbs that form the irregular past tense with the suffix -t.  

In Fig.~\ref{fig:bar_AE_BE}A, we compare American and British English in Ngrams.  The average regularization fraction is 0.49 in AE and $0.42$ in BE.   For 17 out of 22 verbs, AE shows more regularization, with a Wilcoxon signed rank test $p$-value of $9.8\times 10^{-4}$, giving statistical support that AE verbs are more regular on average in Ngrams than BE verbs.  

As we show in the inset scatter plot of Fig.~\ref{fig:bar_AE_BE}A, regularization in AE and BE are also strongly positively correlated with a Spearman correlation coefficient of $0.97$ $(p=2.3\times 10^{-14})$. Verbs that are more regular in AE are also more regular in BE, just not to the same extent.  

In Fig.~\ref{fig:bar_AE_BE}B, we compare regularization in AE and BE on Twitter. For Twitter, the average regularization fraction is $0.54$ for AE, higher than Ngrams, and $0.33$ for BE, much lower than Ngrams.  
As with Ngrams, 17 verbs out of 22 show more regularization in AE than in BE.  
The Wilcoxon signed rank test gives a weaker but still significant $p$-value of $1.9\times 10^{-3}$.  

The inset in Fig.~\ref{fig:bar_AE_BE}B also shows a positive correlation, although not as strong as Ngrams, with a Spearman correlation coefficient of $0.87$ $(p=1.1\times 10^{-7})$. Generally on Twitter, regular AE verbs are also regular in BE, but the difference in regularization fraction is much greater than for Ngrams.  

In Fig.~\ref{fig:bar_diff}A, we demonstrate the difference in regularization between AE and BE for both Ngrams and Twitter. The values in this figure for Ngrams can be thought of as, for each verb in Fig.~\ref{fig:bar_AE_BE}A, subtracting the value of the bottom bar from the top bar, and likewise for Twitter and Fig.~\ref{fig:bar_AE_BE}B.  Positive numbers imply greater regularization in AE, the more common scenario. When the difference is near zero for one corpus, it is usually close to zero for the other corpus as well.  However, when Ngrams shows that AE is notably more regular than BE, Twitter tends to show a much larger difference. 

The average difference in regularization fraction between AE and BE for Twitter is $0.21$, whereas it is only $0.08$ for Ngrams.  Again, we find that these averages are significantly different with a Wilcoxon signed rank $p$-value of $1.9\times 10^{-2}$.  

The inset scatter plot tells a similar story, with a cluster of points near the origin.  As the difference in regularization fraction between regions increases in Ngrams, it also tends to increase in Twitter, with Spearman correlation coefficient $0.65$ and $p$-value $1.0\times 10^{-3}$. The steep rise shows that the difference increases faster on Twitter than in Ngrams.  

Fig.~\ref{fig:bar_diff}B returns to comparing Ngrams and Twitter, but now between AE and BE.  
For each verb, the bar chart shows the difference between the regularization fraction for Twitter and Ngrams in both AE and BE, with positive values showing that regularization for Twitter is greater.  
In this case, the values can be thought of as subtracting the values for the bars in Fig.~\ref{fig:bar_AE_BE}A from the corresponding bars in Fig.~\ref{fig:bar_AE_BE}B.
As we find for English in general, regularization is greater on Twitter than in Ngrams for AE, with an average difference of $0.04$. 
However, for BE, regularization is greater in Ngrams than on Twitter, with an average difference in regularization fraction of $-0.09$.  

\begin{table}[htp!]
\begin{center}
\begin{tabular}{c|c|c||c}
 & Twitter & Ngrams & Difference \\
\hline
AE & 0.54 &  0.49 & 0.04  \\
\hline
BE & 0.33 &  0.42 & $-0.09$ \\
\hline \hline
Difference & $0.21$ & $0.08$ & \\
\end{tabular}
\end{center}
\caption{A summary of the average regularization fractions for AE and BE on Twitter and Ngrams.  Note that the differences were taken prior to rounding.  }\label{table:BEvAE}
\end{table}

\begin{figure}[tp!]
\begin{center}
\includegraphics[width=.43\textwidth]{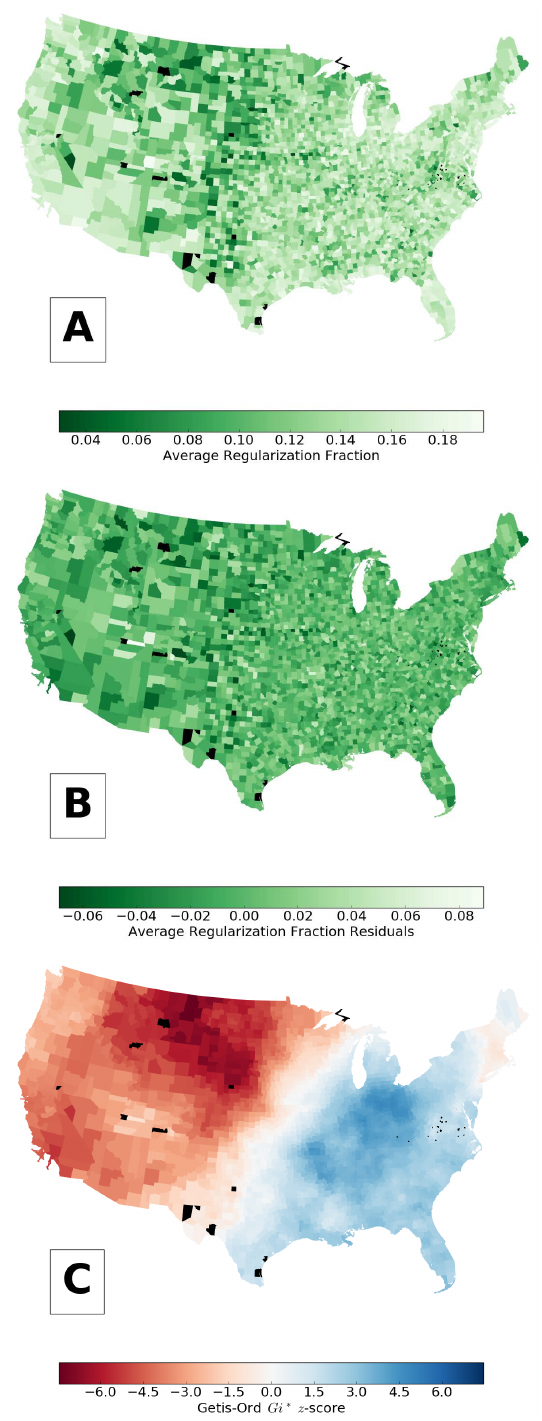}
\end{center}
\vspace{-12pt}
\caption{(A) The average verb regularization fraction by county for the lower 48 states, along with (B) residuals and (C) $Gi^*$ $z$-score.  A higher $Gi^*$ $z$-score means a county has a greater regularization fraction than expected.  Counties colored black did not have enough data.  We used the dataset in row (IV) of Table~\ref{table:datasets}. }\label{regMap}
\end{figure}

We summarize our findings in Table~\ref{table:BEvAE}. We found again that verbs on Twitter are more regular than in Ngrams for American English, likely for many of the same reasons that verbs on Twitter are more regular than Ngrams in general.  However, we find that in British English the opposite is true: Verbs on Twitter are less regular than in Ngrams. In decreasing order by average regularization fraction, we have AE Twitter, then AE Ngrams, then BE Ngrams, and finally BE Twitter.  Knowing that the general trend is towards regularization \cite{ngrams,celex}, it seems that regularization is perhaps being led by everyday speakers of American English, with American published work following suit, but with a lag.  Then, it may be that British English authors and editors are being influenced by American publications and the language used therein. Indeed, some studies have found a general `Americanization' of English across the globe \cite{americanization, baker_2017}, meaning that the various varieties of English used across the world are becoming more aligned with American English.  Finally, it may be that average British users of Twitter are more resistant to the change.  Indeed, from the figures in the study by Gon\c{c}alves et al.,\ one can see that the `Americanization' of British English is more pronounced in Ngrams than on Twitter \cite{americanization}, agreeing with what we have found here.  

\subsection{Regularization by US county}\label{sec:county}

In Sec.~\ref{sec-BEvAE}, we demonstrated regional differences in verb regularization by comparing BE and AE. Here, we consider differences on a smaller spatial scale by quantifying regularization by county in the United States using the dataset in row (IV) of Table~\ref{table:datasets}.  We use methods inspired by Grieve et al.\ to study regional variation in language \cite{grieve_stats}.

We only include counties that had at least 40 total tokens for the verbs under consideration. We plot the average regularization fraction for each county in the continental U.S. in Fig.~\ref{regMap}A, where counties with not enough data are colored black.  To control for the skewed distribution of samples associated with county population (see below for more details), we use residuals for this portion of the analysis. After regressing with the $\log_{10}$ of data volume (total number of tokens) for each county, we compute the average regularization fraction residual, which is plotted in Fig.~\ref{regMap}B.

That is, if we let $d_i$ be the total number of tokens for verbs in tweets from county $i$; $\alpha$ and $\beta$ be the slope and intercept parameters computed from regression; and $R_i$ be the average regularization fraction for county $i$, then we compute the average regularization fraction residual for county $i$, $r_i^{\text{reg}}$, as
\begin{equation}\label{equation:reg_residual}
r_i^{\text{reg}} = R_i - \left(\beta + \alpha \log_{10} d_i  \right).
\end{equation}

Using the average regularization residual at the county level as input, we measure local spatial autocorrelation using the Getis-Ord $Gi^*$ $z$-score \cite{getis-ord},
\begin{equation}\label{equation:gi}
G_i^* = 
\frac{ \sum_j w_{ij} r_j^{\text{reg}} - \overline{r}^{\text{reg}}\sum_j w_{ij}}
{\sigma\sqrt{\left[n\sum_j w_{ij}^2 - \left( \sum_j w_{ij}\right)^2 \right]/(n-1)}},
\end{equation}
where
\begin{equation}
\sigma=\sqrt{
\frac{\sum_j (r_j^{\text{reg}})^2 }{n}
- (\overline{r}^{\text{reg}})^2
},
\end{equation}
 $\overline{r}^{\text{reg}} = \frac{1}{n}\sum_i r_i^{\text{reg}}$, $n$ is the number of counties, and $w_{ij}$ is a weight matrix.
  To obtain the weight matrix used in this calculation, we first create a distance matrix, $s_{ij}$, where the distance between each pair of counties is the larger of the great circle distance, $s_{ij}^\text{GC}$, in miles between the centers of the bounding box for each county and 10 miles. That is,
\begin{equation}
s_{ij}=\max\left(s_{ij}^\text{GC}, 10\right).
\end{equation}
We make the minimum value for $s_{ij}$ 10 miles to prevent a county from having too large of a weight. We then compute the weight matrix as
\begin{equation}
w_{ij}=\frac{1}{\sqrt{s_{ij}}}.
\end{equation}

Fig.~\ref{regMap}C shows the results for the lower 48 states, where black represents counties left out because there was not enough data.  For each county, the $Gi^*$ $z$-score computes a local weighted sum of the residuals, $r_j^\text{reg}$, for the surrounding counties and compares that to the expected value of that weighted sum if all the counties had exactly the average residual, $\overline{r}^\text{reg}$, as their value, where the weighting is such that closer counties have a higher weight.  Areas that are darker blue (positive $z$-score) belong to a cluster of counties that has higher regularization than average, and those that are darker red (negative $z$-score) belong to a cluster that has lower regularization than average.  So, Fig.~\ref{regMap}C shows that, in general, western counties show less regularization than average and eastern counties show more, except that the New England area is fairly neutral.  

\begin{figure}[tp!]
\begin{center}
\includegraphics[width=.5\textwidth]{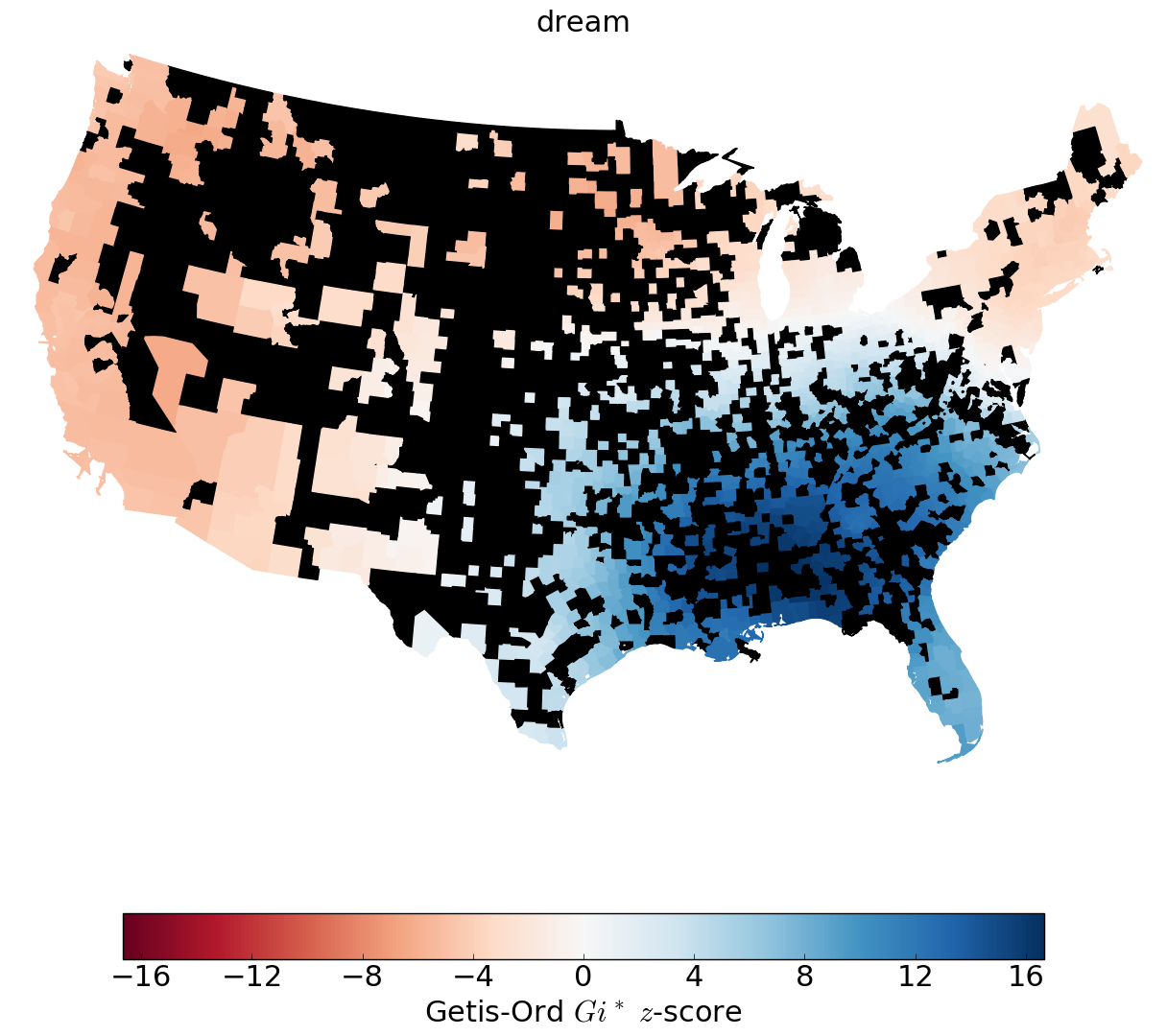}
\end{center}
\caption{ The $Gi^*$ $z$-score for verb regularization by county for the verb `dream' for the lower 48 states.  Counties colored black did not have enough data. People tweet `dreamed' rather than `dreamt' more often than expected in the southeastern U.S.}\label{dreamMap}
\end{figure}

\begin{figure}[tp!]
\begin{center}
\includegraphics[width=.5\textwidth]{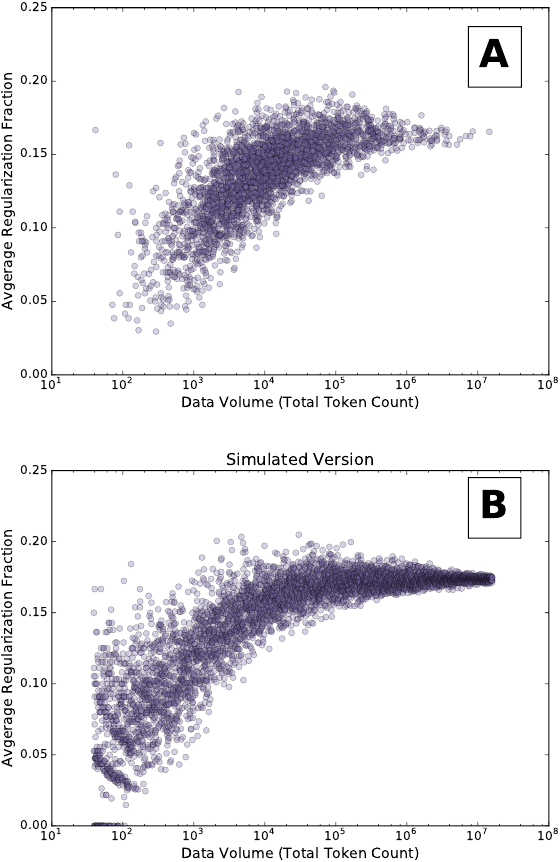}
\end{center}
\caption{(A) Scatter plot of average verb regularization for counties. For each county, the horizontal coordinate is the total token count of verbs found in tweets from that county, and the vertical coordinate is that county's average regularization fraction. For a version with verbs split into frequency bins, see 
\revtexlatexswitch{Fig.~\ref{figure:binned_reg} in Appendix \ref{appendix:binned_reg}.}{S1 Fig.}
  (B) We created synthetic counties by sampling words from the collection of all occurrences of all verbs on Twitter (using the dataset from row (I) of Table~\ref{table:datasets}). The point's horizontal position is given by the total sample token count in a synthetic county; the vertical position is given by its average regularization fraction.  }\label{regularizationVdata}
\end{figure}

As usual, the $z$-score gives the number of standard deviations away from the mean.  For this we would do a two tail test for significance because we are looking for both high value and low value clusters.  For example, a $z$-score greater in magnitude than $1.96$ is significant at the $.05$ level.  If we do a Bonferroni correction based on 3161 counties (the number included for this part of the analysis), then a $z$-score greater in magnitude than $4.32$ is significant for a two tail test at the $.05/3161\approx 1.58 \times 10^{-5}$ level.  

We do this same process looking at individual verbs as well.  However, when looking at individual verbs, we use the regularization fraction rather than residuals, because the data skew is not as problematic.  This is because the main problem with data volume comes when averaging across verbs that have different frequencies of usage, as explained below.  
Also, here we include counties that have at least 10 tokens.  
Fig.~\ref{dreamMap} gives an example map showing the $Gi^*$ $z$-scores for the verb `dream'.  The maps showing local spatial autocorrelation for the complete list of verbs can be found in the Online Appendix A at \onlineappendixurl.  

For many of the counties in the US, there is a small sample of Twitter data. We restrict our analysis to counties with a total token count of at least 40 for the verbs we consider. Even for the counties meeting this criteria, the volume of data varies, leading to drastically different sample sizes across counties.  

More common verbs tend to have popular irregular forms (e.g., `found' and `won'), and less common verbs tend to be regular (e.g., `blessed' and `climbed') \cite{celex}. As a result, samples taken from populous counties are more likely to contain less common verbs.
Our `average regularization' is an average of averages, resulting in an underlying trend toward higher rates for more populous counties due to the increased presence of rarer regular verbs. 

Fig.~\ref{regularizationVdata} demonstrates the relationship between data volume and regularization. To explore the connection further, we perform a synthetic experiment as follows. 

To simulate sampling from counties with varying population sizes, we first combine all verb token counts (using the Twitter dataset from row (I) of Table~\ref{table:datasets}) into a single collection. We then randomly sample a synthetic county worth of tokens from this collection. For a set of 1000 logarithmically spaced county sizes, we randomly draw five synthetic collections of verbs (each is a blue circle in Fig.~\ref{regularizationVdata}). For each sample, we compute the average regularization fraction, as we did for U.S. counties. The goal is to infer the existences of any spurious trend introduced by the sampling of sparsely observed counties.

The resulting simulated curve is comparable to the trend observed for actual U.S. counties. As the data volume increases, the simulated version converges on roughly $0.17$, which is the average regularization fraction for all of Twitter. 

\begin{figure}[tp!]
\begin{center}
\includegraphics[width=.5\textwidth]{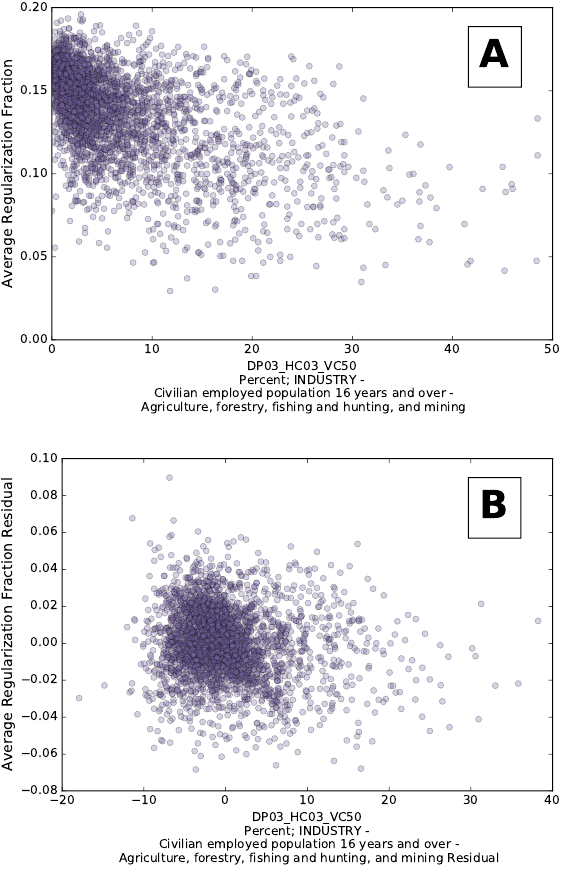}
\end{center}
\caption{(A) Average verb regularization for counties as a function of the percentage of civilians employed in agriculture, forestry, fishing, hunting, and mining. Several hundred such plots are available in an interactive online appendix. (B) For each county, the horizontal coordinate is given by the residual left after regressing the demographic variable with the $\log_{10}$ of data volume and the vertical coordinate is given by the residual left after regressing that county's average regularization fraction with the $\log_{10}$ of data volume.  Data volume, for a county, is the total token count of all verbs found in tweets from that county. 
}\label{partialCorr}
\end{figure}

We also explored correlations between verb regularization and various demographic variables. Fig.~\ref{regularizationVdata} showed a strong relationship between data volume and verb regularization.  It has been shown elsewhere that tweet density positively correlates with population density \cite{populationVStweets}, and population size is correlated with many demographic variables. As a result, we use partial correlations as an attempt to control for the likely confounding effect of data volume.  

For each demographic variable, we compute the regression line between the $\log_{10}$ of data volume, $d_i$, and regularization, and compute the residuals as in Eq.~\ref{equation:reg_residual}.  Then, if the demographic variable is an `Estimate' variable, where the unit is number of people, we similarly compute the regression line between the $\log_{10}$ of data volume and the $\log_{10}$ of the demographic variable\revtexlatexswitch{\footnote{
We do not include any county that has a value of zero for the demographic variable here to prevent errors when taking the $\log_{10}$.}}{ (for all counties where the value of this variable is nonzero)} 
and compute the residuals, $r_i^{\text{dem}}$, as
\begin{equation}
r_i^{\text{dem}} = \log_{10}(D_i) - \left( \delta + \gamma \log_{10} d_i  \right),
\end{equation}
where $D_i$ is the value of the demographic variable for county $i$, and $\gamma$ and $\delta$ are the slope and intercept parameters calculated during regression.  

\begin{table*}[htp!]
\revtexlatexswitch{}{\begin{adjustwidth}{-2.25in}{0in}}
\begin{center}
\begin{tabular}{|c|S|C{11.5cm}|}
\hline
Rank & {Partial Correlation} & Demographic Variable \\
\hline
1 & -0.18 & Percent; OCCUPATION - Civilian employed population 16 years and over - Management, business, science, and arts occupations \\ 
\hline 
2 & -0.16 & Percent; UNITS IN STRUCTURE - Total housing units - 10 to 19 units \\ 
\hline 
3 & -0.16 & Percent; CLASS OF WORKER - Civilian employed population 16 years and over - Self-employed in own not incorporated business workers \\ 
\hline 
4 & -0.16 & Percent; UNITS IN STRUCTURE - Total housing units - 20 or more units \\ 
\hline 
5 & 0.16 & Percent; COMMUTING TO WORK - Workers 16 years and over - Car, truck, or van -- drove alone \\ 
\hline 
6 & 0.15 & Percent; BEDROOMS - Total housing units - 3 bedrooms \\ 
\hline 
7 & -0.15 & Percent; COMMUTING TO WORK - Workers 16 years and over - Worked at home \\ 
\hline 
8 & -0.15 & Percent; INDUSTRY - Civilian employed population 16 years and over - Agriculture, forestry, fishing and hunting, and mining \\ 
\hline 
9 & -0.15 & Percent; BEDROOMS - Total housing units - 1 bedroom \\ 
\hline 
10 & 0.14 & Percent; OCCUPATION - Civilian employed population 16 years and over - Production, transportation, and material moving occupations\\
\hline
\end{tabular}
\end{center}
\caption{Top demographic variables sorted by the magnitude of their partial correlation with verb regularization in U.S. counties. For example, regularization is positively correlated with the percentage of workers driving alone to work, and anti-correlated with the percentage of individuals working from home. Statistics for all of the demographic variables can be found in the Online Appendix B at \onlineappendixurl.}\label{table:partial_correlation}
\revtexlatexswitch{}{\end{adjustwidth}}
\end{table*}

Otherwise, the demographic variable is a `Percent' variable, with units of percentage, and we compute the regression line between the $\log_{10}$ of data volume and the demographic variable, and compute residuals as 
\begin{equation}
r_i^{\text{dem}} = D_i - \left( \delta + \gamma \log_{10} d_i \right).
\end{equation}
The correlation between residuals $r_i^{\text{reg}}$ and $r_i^{\text{dem}}$ gives the partial correlation between average regularization and the demographic variable.  

Our findings suggest that data volume is a confounding variable in at least some of the cases because, after controlling for data volume, there is generally a large decrease in the correlation between verb regularization and the demographic variables. The largest in magnitude Pearson correlation between verb regularization and a demographic variable is $0.68$, for the variable `Estimate; SCHOOL ENROLLMENT - Population 3 years and over enrolled in school', whereas the largest in magnitude partial correlation is only $-0.18$, for the variable `Percent; OCCUPATION - Civilian employed population 16 years and over - Management, business, science, and arts occupations'.  Table~\ref{table:partial_correlation} lists the 10 demographic variables with largest in magnitude partial correlation. 

Fig.~\ref{partialCorr} shows an example for one of the demographic variables, the `Percent' variable with largest simple correlation.  Fig.~\ref{partialCorr}A is the scatter plot of the demographic variable with average regularization, which corresponds to simple correlation.  Fig.~\ref{partialCorr}B is the scatter plot of the residuals, $r_i^{\text{dem}}$ and $r_i^{\text{reg}}$, after regressing with the $\log_{10}$ of data volume, and corresponds with partial correlation.  We can see that there is a strong simple correlation ($-0.52$), but after accounting for data volume that correlation largely vanishes ($-0.15$).  
Similar plots for all of the demographic variables can be found in the Online Appendix B at \onlineappendixurl.

\section{Concluding remarks}
\label{sec:ConcludingRemarks}

Our findings suggest that, by and large, verb regularization patterns are similar when computed with Ngrams and Twitter. However, for some verbs, the extent of regularization can be quite different. If social media is an indicator of changing patterns in language use, Ngrams data ought to lag with a timescale not yet observable due to the recency of Twitter data. Very reasonably, Ngrams data may not yet be showing some of the regularization that is happening in everyday English.  

We also found differences in verb regularization between American and British English, but found that this difference is much larger on Twitter than Ngrams. Overall, and in American English specifically, verbs are more regular on Twitter than in Ngrams, but the opposite is true for British English. 
In the U.S., we also find variation in average verb regularization across counties.  Lastly, we showed that there are significant partial correlations between verb regularization and various demographic variables, but they tend to be weak.  

Our findings do not account for the possible effects of spell checkers.  Some people, when tweeting, may be using a spell checker to edit their tweet.  If anything, this will likely skew the language on Twitter towards the `correct' form used in edited textual sources.  For example, in Fig.~\ref{ngramsVtwitter} we see that `stand' is irregular for both Ngrams and Twitter, and likely most spell checkers would consider the regular `standed' a mistake, but we see that `stand' is still over 100 times more regular on Twitter than in Ngrams.   So, the differences between edited language and everyday language may be even larger than what we find here suggests.  Future work should look into the effects of spell checkers.  

Our study explored the idea that edited written language may not fully represent the language spoken by average speakers.  However, tweets do not, of course, fully represent the English speaking population. Even amongst users, our sampling is not uniform as it reflects the frequency with which different users tweet 
\revtexlatexswitch{(see Fig.~\ref{figure:user_frequency_plot} in Appendix \ref{appendix:user_frequency_plot})}{(see S2 Fig.)}.  
Furthermore, the language used on Twitter is not an unbiased sample of language even for people who use it frequently. The way someone spells a word and the way someone pronounces a word may be different, especially, for example, the verbs with an irregular form ending in -t, because -t and -ed are close phonetically. However, the fact that we found differences between the language of Ngrams and the language of Twitter suggests that the true language of everyday people is not fully represented by edited written language.  We recommend that future studies should investigate speech data.

\acknowledgments
We are thankful for the helpful reviews and discussions of earlier versions of this work by A. Albright and J. Bagrow, and for help with initial data collection from L. Gray. PSD \& CMD were supported by NSF Grant No. IIS-1447634, and TJG, PSD, \& CMD were supported by a gift from MassMutual.

\bibliographystyle{unsrt}

\begin{thebibliography}{10}

\bibitem{celex}
Erez Lieberman, Jean-Baptiste Michel, Joe Jackson, Tina Tang, and Martin~A
  Nowak.
\newblock Quantifying the evolutionary dynamics of language.
\newblock {\em Nature}, 449(7163):713--716, 10 2007.

\bibitem{ngrams}
Jean-Baptiste Michel, Yuan~Kui Shen, Aviva~Presser Aiden, Adrian Veres,
  Matthew~K. Gray, {\relax The Google Books Team}, Joseph~P. Pickett, Dale
  Hoiberg, Dan Clancy, Peter Norvig, Jon Orwant, Steven Pinker, Martin~A.
  Nowak, and Erez~Lieberman Aiden.
\newblock Quantitative analysis of culture using millions of digitized books.
\newblock {\em Science}, 331(6014):176--182, 2011.

\bibitem{nature}
Mitchell~G. Newberry, Christopher~A. Ahern, Robin Clark, and Joshua~B. Plotkin.
\newblock Detecting evolutionary forces in language change.
\newblock {\em Nature}, 551(7679):223, 2017.

\bibitem{GreenhillE8822}
Simon~J. Greenhill, Chieh-Hsi Wu, Xia Hua, Michael Dunn, Stephen~C. Levinson,
  and Russell~D. Gray.
\newblock Evolutionary dynamics of language systems.
\newblock {\em Proceedings of the National Academy of Sciences},
  114(42):E8822--E8829, 2017.

\bibitem{Reali20172586}
Florencia Reali, Nick Chater, and Morten~H. Christiansen.
\newblock Simpler grammar, larger vocabulary: How population size affects
  language.
\newblock {\em Proceedings of the Royal Society of London B: Biological
  Sciences}, 285(1871), 2018.

\bibitem{Ramiro201714730}
Christian Ramiro, Mahesh Srinivasan, Barbara~C. Malt, and Yang Xu.
\newblock Algorithms in the historical emergence of word senses.
\newblock {\em Proceedings of the National Academy of Sciences},
  115(10):2323--2328, 2018.

\bibitem{dodds}
Eitan~Adam Pechenick, Christopher~M. Danforth, and Peter~Sheridan Dodds.
\newblock Characterizing the {G}oogle {B}ooks corpus: {S}trong limits to
  inferences of socio-cultural and linguistic evolution.
\newblock {\em PLOS ONE}, 10(10):1--24, 10 2015.

\bibitem{eitan2}
Eitan~Adam Pechenick, Christopher~M. Danforth, and Peter~Sheridan Dodds.
\newblock Is language evolution grinding to a halt? {T}he scaling of lexical
  turbulence in {E}nglish fiction suggests it is not.
\newblock {\em Journal of Computational Science}, 21:24--37, 2017.

\bibitem{pew}
The demographics of social media users.
\newblock
  \url{http://www.pewinternet.org/2015/08/19/the-demographics-of-social-media-users/},
  2015.
\newblock Accessed: 2016-07-18.

\bibitem{ngramsViewer}
{\relax The Google Ngram Viewer Team}.
\newblock Google {N}gram {V}iewer.
\newblock \url{https://books.google.com/ngrams/}, 2013.

\bibitem{ngramsInfo}
{\relax The Google Ngram Viewer Team}.
\newblock About {N}gram {V}iewer.
\newblock \url{https://books.google.com/ngrams/info}, 2013.

\bibitem{americanization}
Bruno Gon{\c{c}}alves, Luc{\'{\i}}a Loureiro{-}Porto, Jos{\'{e}}~J. Ramasco,
  and David S{\'{a}}nchez.
\newblock Mapping the {A}mericanization of {E}nglish in space and time.
\newblock {\em PLOS ONE}, 13(5):1--15, 05 2018.

\bibitem{baker_2017}
Paul Baker.
\newblock {\em American and British English: Divided by a Common Language?}
\newblock Cambridge University Press, 2017.

\bibitem{grieve_stats}
Jack Grieve, Dirk Speelman, and Dirk Geeraerts.
\newblock A statistical method for the identification and aggregation of
  regional linguistic variation.
\newblock {\em Language Variation and Change}, 23(2):193--221, 2011.

\bibitem{getis-ord}
J.~K. Ord and Arthur Getis.
\newblock Local spatial autocorrelation statistics: Distributional issues and
  an application.
\newblock {\em Geographical Analysis}, 27(4):286--306, 1995.

\bibitem{populationVStweets}
Rudy Arthur and Hywel T.~P. Williams.
\newblock Scaling laws in geo-located {T}witter data.
\newblock {\em CoRR}, abs/1711.09700, 2017.

\end{thebibliography}

\clearpage

\onecolumngrid
\appendix

\setcounter{table}{0}
\renewcommand{\thetable}{\Alph{section}\arabic{table}}
\setcounter{figure}{0}
\renewcommand{\thefigure}{\Alph{section}\arabic{figure}}

\section{Table of Verb Forms}\label{appendix:verb_table}

\begin{longtable*}{|c||c||c|c||r|}
\hline
 & Regular & \multicolumn{2}{c||}{Irregular} & \\
Verb & Preterit \& Past Participle & Preterit & Past Participle & Token Count\\
\hline
\endfirsthead

\caption{(continued)}\\
\hline
 & Regular & \multicolumn{2}{c||}{Irregular} & \\
Verb & Preterit \& Past Participle & Preterit & Past Participle & Token Count\\
\hline
\endhead

\hline
\multicolumn{5}{|r|}{Continued on next page} \\
\hline 
\caption{
A tabulation of all verb forms used in this study.  The Token Count column gives the sum of all the tokens for the past tense forms of the verb, both regular and irregular, in our Twitter dataset (see row (I) of Table~\ref{table:datasets} in Sec.~\ref{sec:data}).
}\label{table:verb_table}\\
\endfoot

\hline
\endlastfoot

abide & abided & abode & abode & 146,566 \\ 
alight & alighted & alit & alit & 56,306 \\ 
arise & arised & arose & arisen & 164,134 \\ 
awake & awaked & awoke & awoken, awoke & 423,359 \\ 
become & becomed & became & become & 50,664,026 \\ 
begin & beginned & began & begun & 5,942,572 \\ 
bend & bended & bent & bent & 4,777,019 \\ 
beseech & beseeched & besought & besought & 3,390 \\ 
bleed & bleeded & bled & bled & 252,225 \\ 
blend & blended & blent & blent & 436,029 \\ 
bless & blessed & blest & blest & 22,547,387 \\ 
blow & blowed & blew & blown & 9,155,246 \\ 
break & breaked & broke & broken & 54,506,810 \\ 
breed & breeded & bred & bred & 1,040,854 \\ 
bring & bringed & brought & brought & 15,303,318 \\ 
build & builded & built & built & 8,521,553 \\ 
burn & burned & burnt & burnt & 7,457,942 \\ 
buy & buyed & bought & bought & 24,841,526 \\ 
catch & catched & caught & caught & 24,891,188 \\ 
choose & choosed & chose & chosen & 10,290,205 \\ 
clap & clapped & clapt & clapt & 405,837 \\ 
climb & climbed & clomb, clom & clomben & 635,122 \\ 
cling & clinged & clung & clung & 49,742 \\ 
creep & creeped & crept & crept & 698,405 \\ 
deal & dealed & dealt & dealt & 1,181,974 \\ 
dig & digged & dug & dug & 941,656 \\ 
dream & dreamed & dreamt & dreamt & 2,794,060 \\ 
drink & drinked & drank & drunk, drank & 37,295,703 \\ 
drive & drived & drove & driven & 5,745,497 \\ 
dwell & dwelled & dwelt & dwelt & 25,725 \\ 
eat & eated & ate & eaten & 25,084,758 \\ 
fall & falled & fell & fallen & 25,224,815 \\ 
fight & fighted & fought & fought & 3,625,297 \\ 
find & finded & found & found & 80,709,195 \\ 
flee & fleed & fled & fled & 405,295 \\ 
freeze & freezed & froze & frozen & 7,454,847 \\ 
get & getted & got & got, gotten & 500,591,203 \\ 
give & gived & gave & given & 58,697,198 \\ 
grow & growed & grew & grown & 17,951,273 \\ 
hang & hanged & hung & hung & 3,991,956 \\ 
hear & heared & heard & heard & 52,605,822 \\ 
hide & hided, hidded & hid & hid, hidden & 7,829,276 \\ 
hold & holded & held & held & 10,080,725 \\ 
inlay & inlayed & inlaid & inlaid & 44,811 \\ 
keep & keeped & kept & kept & 11,785,131 \\ 
kneel & kneeled & knelt & knelt & 83,765 \\ 
know & knowed & knew & known & 58,175,701 \\ 
lay & layed & laid & laid & 5,828,898 \\ 
leap & leaped & leapt & leapt & 91,620 \\ 
learn & learned & learnt & learnt & 18,134,586 \\ 
lose & losed & lost & lost & 72,695,892 \\ 
mean & meaned & meant & meant & 26,814,977 \\ 
pay & payed & paid & paid & 21,150,031 \\ 
plead & pleaded & pled & pled & 193,553 \\ 
ride & rided & rode & ridden & 1,710,109 \\ 
seek & seeked & sought & sought & 888,822 \\ 
sell & selled & sold & sold & 14,251,612 \\ 
send & sended & sent & sent & 26,265,441 \\ 
shake & shaked & shook & shaken & 3,223,316 \\ 
shoe & shoed & shod & shod & 47,780 \\ 
shrink & shrinked & shrank, shrunk & shrunk, shrunken & 296,188 \\ 
sing & singed & sang, sung & sung & 6,767,707 \\ 
sink & sinked & sank, sunk & sunk, sunken & 927,419 \\ 
slay & slayed & slew & slain & 2,153,981 \\ 
sleep & sleeped & slept & slept & 9,252,446 \\ 
slide & slided & slid & slid & 530,659 \\ 
sling & slinged & slung & slung & 38,320 \\ 
slink & slinked & slunk & slunk & 5,772 \\ 
smell & smelled & smelt & smelt & 1,089,814 \\ 
smite & smitted, smited & smote & smitten, smote & 176,768 \\ 
sneak & sneaked & snuck & snuck & 797,337 \\ 
speak & speaked & spoke & spoken & 8,502,050 \\ 
speed & speeded & sped & sped & 216,062 \\ 
spell & spelled & spelt & spelt & 3,812,137 \\ 
spend & spended & spent & spent & 17,603,781 \\ 
spill & spilled & spilt & spilt & 1,627,331 \\ 
spin & spinned & spun & spun & 342,022 \\ 
spoil & spoiled & spoilt & spoilt & 3,891,576 \\ 
spring & springed & sprang, sprung & sprung & 626,400 \\ 
stand & standed & stood & stood & 3,942,812 \\ 
steal & stealed & stole & stolen & 11,884,934 \\ 
sting & stinged & stung & stung & 391,053 \\ 
stink & stinked & stank, stunk & stunk & 1,556,197 \\ 
stride & strided & strode & stridden & 17,811 \\ 
strike & striked & struck & struck, stricken & 2,167,165 \\ 
strip & stripped & stript & stript & 837,967 \\ 
strive & strived & strove & striven & 33,705 \\ 
swear & sweared & swore & sworn & 1,902,662 \\ 
sweep & sweeped & swept & swept & 931,245 \\ 
swim & swimmed & swam & swum & 356,842 \\ 
swing & swinged & swung & swung & 295,360 \\ 
take & taked & took & taken & 83,457,822 \\ 
teach & teached & taught & taught & 9,379,039 \\ 
tear & teared & tore & torn & 4,238,865 \\ 
tell & telled & told & told & 71,562,969 \\ 
thrive & thrived & throve & thriven & 43,612 \\ 
throw & throwed & threw & thrown & 13,197,226 \\ 
tread & treaded & trod & trodden & 56,624 \\ 
vex & vexed & vext & vext & 139,411 \\ 
wake & waked & woke & woken & 30,796,918 \\ 
wear & weared & wore & worn & 8,552,191 \\ 
weep & weeped & wept & wept & 200,690 \\ 
win & winned & won & won & 45,276,202 \\ 
wind & winded & wound & wound & 1,340,267 \\ 
wring & wringed & wrung & wrung & 29,141 \\ 
write & writed & wrote & written, writ, wrote & $23,926,025$\\
\end{longtable*}

\newpage

\section{Details on User Location Matching}\label{appendix:location_match_info}
\twocolumngrid
To study regularization by county, we extracted location information from the user-provided location information, which was entered as free text in the user's biographical profile.  To do this, for each tweet we first checked if the location field was populated with text.  If so, we then split the text on commas, and checked whether there were two tokens separated by a comma.  If so, we made the assumption that it might be of the form `city, state'.  Then we used a python package called uszipcode, which can be found here:  
\url{pythonhosted.org/uszipcode/}.
We used the package's method to search by city and state.  If the package returned a location match, we used the returned latitude and longitude to determine which county the detected city belonged to.  

The package allows for fuzzy matching, meaning the city and state do not have to be spelled correctly, and it allows for the state to be fully spelled out or be an abbreviation.  In the source code of the package there was a hard coded confidence level of 70 for the fuzzy matching.  We modified the source code so that the confidence level was an input to the method, and running tests found we were satisfied with a confidence level of 91.  We checked by hand the matches of 1000 tweets that this method returned a match for, 100 from each year in the dataset, and found the only potential error in these matches was when the user typed in `Long Island, NY', or a similar variant.  For this, the package returned Long Island City, NY, which is on Long Island, but there are multiple counties on Long Island, so the user may actually live in a different county.  None of the other 1000 tweets were inappropriately or ambiguously assigned.

\onecolumngrid

\newpage

\section{}\label{appendix:binned_reg}

\begin{figure}[hbp!]
\begin{center}
\includegraphics[width=\textwidth]{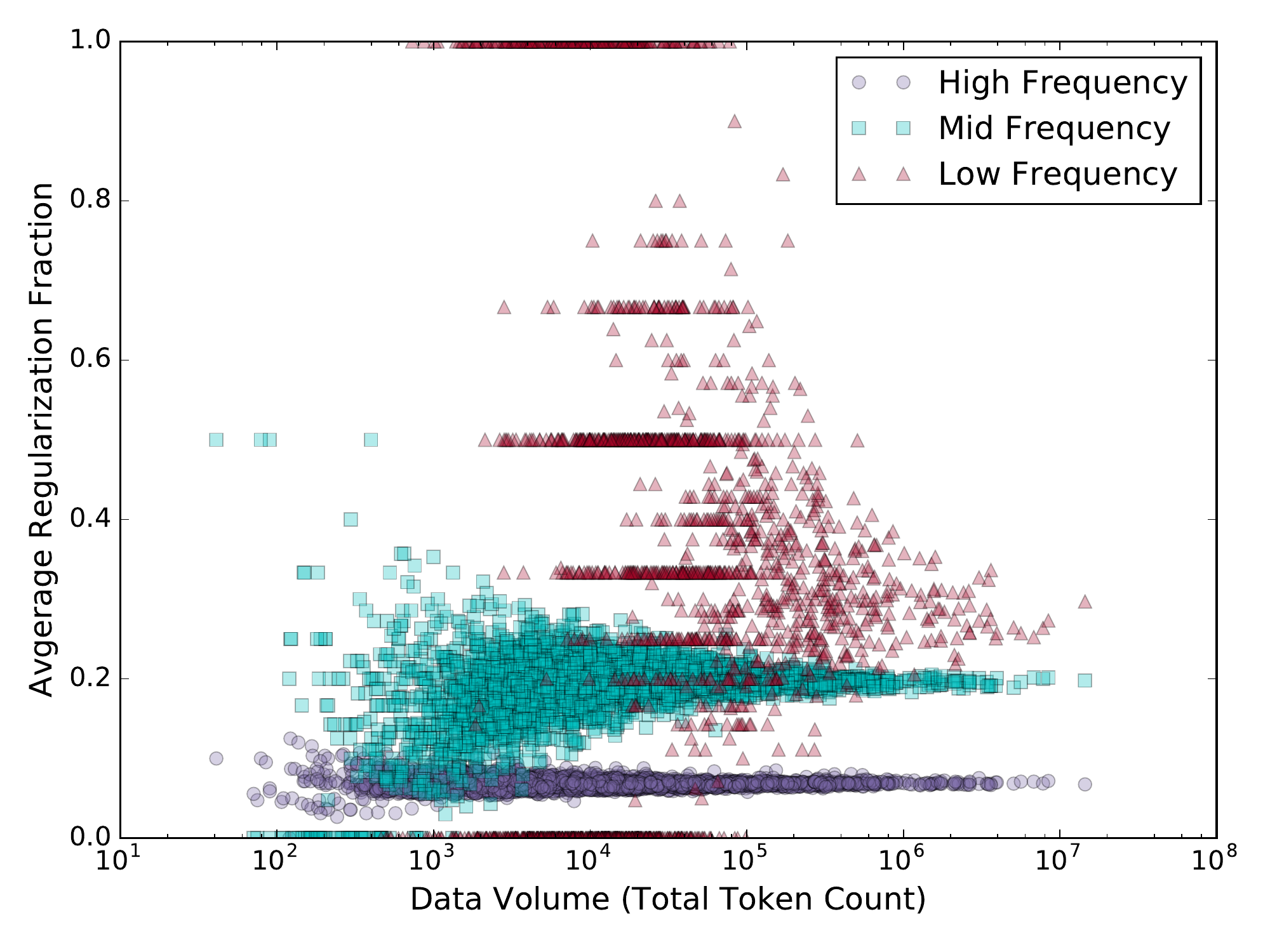}
\end{center}
\caption{The scatter plot of average binned verb regularization for counties.  Verbs with a token count in the interval $[10^6, 10^8]$ in the Twitter dataset from row (IV) of Table~\ref{table:datasets} in Sec.~\ref{sec:data} are considered `high frequency', those in the interval $[10^4, 10^6)$ are `mid frequency', and those in the interval $[10^2, 10^4)$ are low frequency.  The bins contain 37, 55, and 14 verbs, respectively.  For each county (with at least 40 total tokens), the average regularization fraction of the verbs in each of the three bins is calculated (if it is not empty) and plotted against the total token count for all verbs for that county.
}\label{figure:binned_reg}
\end{figure}

\newpage

\section{}\label{appendix:user_frequency_plot}

\setcounter{figure}{0}

\begin{figure}[hbp!]
\begin{center}
\includegraphics[width=\textwidth]{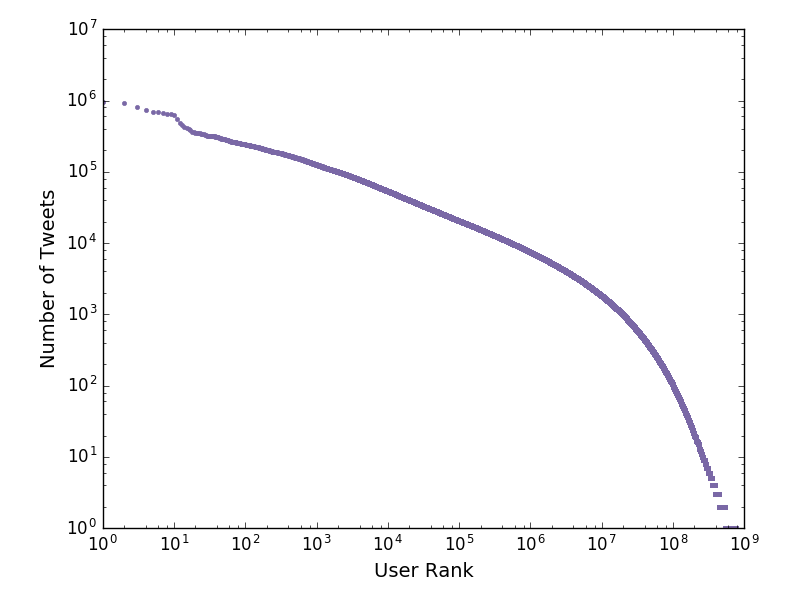}
\end{center}
\caption{The frequency counts of tweets by unique users in our Twitter decahose dataset (row (I) of Table~\ref{table:datasets} in Sec.~\ref{sec:data}).  Users are ranked by their total number of tweets along the horizontal axis and the vertical axis gives the total number of tweets we have associated with each user's account.  
}\label{figure:user_frequency_plot}
\end{figure}

\end{document}